\documentclass[10pt,twocolumn,letterpaper]{article}

\usepackage[pagenumbers]{cvpr} % To force page numbers, e.g. for an arXiv version

\usepackage{graphicx}
\usepackage{amsmath}
\usepackage{amssymb}
\usepackage{booktabs}

\usepackage[pagebackref,breaklinks,colorlinks]{hyperref}

\usepackage[capitalize]{cleveref}
\crefname{section}{Sec.}{Secs.}
\Crefname{section}{Section}{Sections}
\Crefname{table}{Table}{Tables}
\crefname{table}{Tab.}{Tabs.}

\def\cvprPaperID{*****} % *** Enter the CVPR Paper ID here
\def\confName{CVPR}
\def\confYear{2022}

\begin{document}

\title{Asymmetric Graph Representation Learning}

% \author{Zhuo~Tan\\
% Center of Statistical Research\\
% Southwestern University of Finance and Economics\\
% {\tt\small tanzhuo@smail.swufe.edu.cn}
% \and
% Bin~Liu\\
% Center of Statistical Research\\
% Southwestern University of Finance and Economics\\
% % First line of institution2 address\\
% {\tt\small liubin@swufe.edu.cn}
% \and
% Guosheng Yin\\
% Department of Statistics and Actuarial Science\\
% University of Hong Kong\\
% {\tt\small gyin@hku.hk}
% }
\author{
%     %Authors
%     % All authors must be in the same font size and format.
    Zhuo~Tan\textsuperscript{\rm 1},
    Bin~Liu\textsuperscript{\rm 1},
    Guosheng Yin\textsuperscript{\rm 2}\\

%     %Afiliations
    \textsuperscript{\rm 1}Center of Statistical Research,
    Southwestern University of Finance and Economics\\
    \textsuperscript{\rm 2}Department of Statistics and Actuarial Science,
    University of Hong Kong\\
%     % If you have multiple authors and multiple affiliations
%     % use superscripts in text and roman font to identify them.
%     % For example,

%     % Sunil Issar, \textsuperscript{\rm 2}
%     % J. Scott Penberthy, \textsuperscript{\rm 3}
%     % George Ferguson,\textsuperscript{\rm 4}
%     % Hans Guesgen, \textsuperscript{\rm 5}.
%     % Note that the comma should be placed BEFORE the superscript for optimum readability

%     2275 East Bayshore Road, Suite 160\\
%     Palo Alto, California 94303\\
%     % email address must be in roman text type, not monospace or sans serif
    {\tt\small tanzhuo@smail.swufe.edu.cn, liubin@swufe.edu.cn, gyin@hku.hk}
}
\maketitle

%%%%%%%%% ABSTRACT
\begin{abstract}
Despite the enormous success of graph neural networks (GNNs), most existing GNNs can only be applicable to undirected graphs where relationships among connected nodes are two-way symmetric (i.e., information can be passed back and forth). However, there is a vast amount of applications where the information flow is asymmetric, leading to directed graphs where information can only be passed in one direction. For example, a directed edge indicates that the information can only be conveyed forwardly from the start node to the end node, but not backwardly. To accommodate such an asymmetric structure of directed graphs within the framework of GNNs, we propose a simple yet remarkably effective framework for directed graph analysis to incorporate such one-way information passing.
We define an incoming embedding and an outgoing embedding for each node to model its sending and receiving features respectively. 
We further develop two steps in our directed GNN model with the first
one to aggregate/update the incoming features of nodes and the
second one to aggregate/update the outgoing features. By imposing the two roles for each node, the likelihood of a directed edge can be calculated based on the outgoing embedding of the start node and the incoming embedding of the end node. The log-likelihood of all edges plays a natural role of regularization for the proposed model, which can alleviate the over-smoothing problem of the deep GNNs. Extensive experiments on multiple real-world directed graphs demonstrate outstanding performances of the proposed model in both node-level and graph-level tasks.
\end{abstract}

%%%%%%%%% BODY TEXT
\section{Introduction}
\label{sec:intro}

%1 the popularity of GNN
% In recent years, graphs have achieved empirical successes on representing a wide variety of real-world data. Graph-structured data are ubiquitous across various domains, such as social networks \cite{cho2011friendship}, citation networks \cite{kipf2017semi}, quantum chemistry \cite{li2018adaptive}, and combinatorial optimization \cite{selsam2019learning}. Due to the abundance of graph-structured data, GNNs have attracted considerable attention in the machine learning community, with many successful applications, such as text classification \cite{kipf2017semi, yao2019graph}, graph regression \cite{gilmer2017neural, shi2019skeleton} and traffic prediction \cite{song2020spatial}, etc.

GNNs provide a powerful framework for graph representation learning.
%These models offer a powerful tool for performing machine learning on graphs. 
One of the main reasons for GNNs' success 
%can achieve such good results 
is the networks' capability of generating representations of nodes that actually depend on the structure of the graph, as well as node feature information. 
%(might have). 
Most GNNs belong to one of the two categories: spectral-based networks \cite{bruna2014spectral, defferrard2016convolutional, kipf2017semi} or spatial-based networks \cite{hamilton2017inductive, xu2019powerful, klicpera2019predict}.
The spectral-based GNNs use adjacency matrices to explore the neighborhood based on spectral graph theory from the perspective of graph signal processing.
The spatial-based approaches, on the other hand, directly perform the convolution in the graph domain by aggregating information of the neighbor nodes.
% , {\color{red}so that %object 
% each node tries to capture rich information from its neighborhood through the edges within the graph.}
Numerous GNN variants have also been proposed, and most of them can be reformulated into a single common framework \cite{gilmer2017neural}, called message passing neural networks (MPNNs). These models employ a message passing procedure to aggregate local information of vertices.

% 2 most GNNs focus on undirected graphs
% 3 the directed applications are universal, list some important examples from different areas; 
Although GNNs have achieved practical successes 
%improvement 
in many aspects, most of the existing methods only focus on undirected graphs. The directed graphs have been largely ignored or been simplified with undirected settings  \cite{hamilton2017inductive}. However, relationships in directed graphs are asymmetric, which contain essential structural information about the graph.  As a special type of graphs, the directed graphs arise broadly
%are yet broadly seen 
across different domains; for instance, the email networks \cite{ma2019spectral}, citation networks \cite{tong2020digraph}, natural language sentences \cite{yao2019graph}, as well as website networks \cite{zhang2021magnet}, etc. These examples all involve directional edges, indicating receiving and sending information in one-way or two-way flow among nodes. 
 
% 4 directed graphs are more challenge than undirected cases
In undirected graphs, relationships among nodes are symmetric.
% {\color{red}and then a node is equivalent for its neighborhoods.}
Directed graphs, in contrast, have asymmetric node relationships. In particular, some nodes can have edges of both in-taking 
%injection 
(receiving information) and out-flowing (sending information) functions, which makes GNNs learn the graph feature much more difficult than the undirected graph setting. It is challenging to model directed graphs using the current framework of GNNs due to the asymmetric structure. How to incorporate directed structural information into the node representation for higher predictive power is an important %and challenging 
task. All the aforementioned methods, whether the spectral-based or spatial-based approaches, cannot be directly applied to directed graphs. For spectral-based methods, they are limited to use undirected graphs as input by definition, and the graph Laplacian needs to be symmetric. Therefore, extending spectral methods to directed graphs is not straightforward because the adjacency matrix is asymmetric and there is no obvious way to define a symmetric Laplacian. For spatial-based approaches, there are natural advantages on extension to directional settings, but most of the existing methods \cite{hamilton2017inductive, wang2021multi, tong2020directed} are not efficient to capture the complex structure of directed graphs. How to extend GNNs from undirected graphs to directed graphs needs to be further explored.

% 5 current solutions of modeling directed graphs with GNNs: 1) convert directed graphs into undirected settings  2) current directed solutions (review some existing directed GNN solutions) 3) others

% 6 the shortages of current solutions: solution 1) has a problem of loss information; 2) limitations on graph representation learning; from the view of the difference between sending/receiving embeddings ;

The existing GNN solutions for directed graphs fall into one of the two categories. A majority of them is to relax directed graphs to undirected ones, i.e., by trivially adding edges to symmetrize the adjacency matrices \cite{monti2018motifnet, tong2020directed, ma2019spectral, tong2020digraph}. The disadvantages of these methods are obvious. Relaxing directional edges to undirected settings would incur loss of information on the asymmetric node relationships, which cannot represent the distinctive directional structure, such as irreversible citations, hyperlinks, mailings, and time-series relationships, etc. For example, in a citation graph, a later published article can cite a previously published article, but not the opposite due to such a unique time-series relationship. If we convert a citation relationship network into an undirected graph,  noise or bias would be introduced inevitably. Although we can represent the original directed graph in another form, such as a temporal graph learned by a combination of recurrent neural networks (RNNs) and GNNs 
\cite{pareja2020evolvegcn}, we still hope to mine more structural features from the original directed graph without adding additional components.
Some other methods \cite{monti2018motifnet, tong2020directed} have to  stipulate learning templates or rules in advance, which is very complicated and difficult to explain, and is not capable of dealing with complex structures beyond their definitions \cite{tong2020digraph}.

There are also other works aiming to learn specific directed structures by using the incident matrix with attributes of edges \cite{shi2019skeleton, fu2020deep}, through defining a complex Hermitian matrix known as the magnetic Laplacian, aggregating directed predecessors information \cite{wang2021multi, thost2021directed} and defining a vector field with the eigenvectors of graph Laplacian \cite{beani2021directional}. In particular, Beani et al.  \cite{beani2021directional} %the authors 
define directed convolution with vector fields on a graph.
However, these methods for directed graphs either lack  interpretability under the framework of the spectral and spatial graph learning \cite{shi2019skeleton, fu2020deep, wang2021multi, thost2021directed} or are faced with scalability problems \cite{beani2021directional}.

% However, there are two major shortcomings with existing solutions. For the common method of transforming directed graphs to undirected, it will be unable to represent the distinctive direction structure, such as irreversible citation, hyperlink, mailing, and time-series relationships. 
% For example, in a citation graph, a later published article can cite a previously published article, but not vice versa. This is a unique time-series relationship. If we convert it to an undirected graph, we lose this part of information. Although we can represent the original directed graph in another form, such as a temporal graph learned by a combination of Recurrent Neural Networks (RNN) and GCN, we still want to mine more structural features from the original directed graph without adding additional components.
% Other methods have to stipulate learning templates or rules in advance, which is very complicated and difficult to explain, and is not capable to deal with complex structures beyond their definitions.

% 7 introduce our method;
Inspired by the co-clustering algorithm for directional community detection \cite{rohe2016co}, we let one node play two different roles in a directed graph, i.e., the incoming-node and outgoing-node, to accommodate the asymmetries of the directed graph. Specifically, within the framework of GNNs, we introduce two types of embedding with an outgoing vector and an incoming vector (of the same dimension), for each node. The incoming embedding aims to capture receiving features and the outgoing embedding aims to capture sending-out features. 
% To accommodate the asymmetries of directed graph within the framework of GNN, we first extend the GNNs to directed graph by introduced two kinds of embedding, sending embedding and receiving embedding, for each node. Our motivations is inspired by the co-clustering algorithm for directional community detection \cite{rohe2016co}, that the receiving vector to capture incoming feature and the sending embedding to capture outgoing feature, for each node. That is, one node plays two different roles in a directed graph, the incoming-node and outgoing-node.
Figure \ref{fig:inoutEmbedding} illustrates an example of such asymmetric node representations, where node $v$ in a directed graph is described by both an incoming vector $\mathbf{r}_v$ and an outgoing vector $\mathbf{s}_v$. Different from traditional GNNs where each node in an undirected graph is represented as a single vector, the asymmetric structure of the directed graph is characterized by both $\mathbf{r}_v$ and $\mathbf{s}_v$. The asymmetric node representations facilitate interpretable edge representations as well. Any edge in a directed graph can be characterized by the outgoing embedding of its source node and the incoming embedding of its target node.

Traditionally, within the framework of MPNNs, GNNs can be generalized as one aggregating step and one subsequent updating step.
With the asymmetric representations, we extend the traditional GNN framework to the directed graph by defining two aggregating/updating steps, with the first one to aggregate/update the receiving features of nodes and the second one to aggregate/update sending-out features. The asymmetrical property of the proposed GNN makes each node in a directed graph exchange (aggregate/update) ``messages" with the neighbors in two different ways: the in-taking information and the out-flowing information. The incoming embedding and outgoing embedding obtained by GNNs can preserve the structural information of the directed graph. Moreover, the likelihood of a directional edge can be modeled by the outgoing embedding of its source node and the incoming embedding of its target node, which is a natural regularization term of the proposed model that can alleviate the issue of over-smoothing. 
%As we know, 
The embeddings of nodes would be similar due to lager receptive fields through message passing as the number of layers increases \cite{xu2018representation}, which is known as over-smoothing problem of GNNs. The proposed regularization term, in contrast, maintains similarity among 1-hop neighbors, which can balance messages over propagation by restraining the receptive field.
% {\color{red} By regularizing the proposed asymmetric GNNs with the link likelihood, the proposed model can maintain the virginity that nodes are similar to their direct neighborhoods.}

% leveraging the receiving embedding and sending embedding for one node to capturing the receiving information and sending information, respectively. 
% Moreover, it is common method to reconstruct certain graph statistics from the node embeddings that are generated by the encoder in the field of node embedding. 
% We look forward to the receiving embedding and sending embedding obtained by GNNs can preserve the structure information of the directed graph, that is there are as similar as possible embeddings among adjacent nodes and as dissimilar a representation as possible. Therefore, a novel regularization term equips with the log-likelihood to encourage community structure and thus penalize non-community structure among the nodes representations, which is contribute to reduce over-smoothing. As the number of layers increasing, the embedding of nodes maybe similar due to lager receptive filed through message passing, but the regularization formulation will maintain the virginity that nodes are just similar to their direct neighborhoods.

% 8 the contributions: 1) a well-designed architecture for directed GNNs; 2) oversmoothing  regularization ;
% 3) experimental studies to support our model
The contributions of this work are as follows:
\begin{itemize}
    \item We propose a novel directed graph neural network, called asymmetric GNN (AGNN), with the incoming and outgoing embeddings to learn the different roles for each node in directed graphs;
    \item A novel regularization term is proposed to alleviate the over-smoothing issue;
    \item Our approach works well on different directed graph datasets in semi-supervised nodes classification and graph regression tasks, and achieves better performance than the state-of-the-art directed GNN methods.
\end{itemize}

% 9 organization of this paper
The rest of the paper is organized as follows. 
Section 2 introduces the related work in GNNs.
Section 3 formulates the basic GNN models for undirected graphs.
The detailed formulation of our proposed AGNN is given in Section 4.
The ablation study and comparisons with the state-of-the-art methods are shown in Section 5.
% Section 6 provides some quantitative results and discussions.
Section 6 concludes the paper with remarks.

%------------------------------------------------------------------------
\section{Related Work}

\subsection{Directed Graph Neural Networks}

A majority of GNNs transform directed graphs to undirected ones by relaxing its direction structure, i.e., trivially adding edges to symmetrize the adjacency matrices.
Monti, Otness, and Bronstein \cite{monti2018motifnet} deal with directed graphs by exploiting local graph motifs \cite{benson2016higher}, which define a new set of symmetric motif adjacency matrices. 
Tong et al. \cite{tong2020directed} use first-order and second-order proximity to obtain symmetric adjacency matrices, which not only can retain the connection properties of the directed graph, but also expand the receptive field of the convolution operation.
Ma et al. \cite{ma2019spectral} and Tong et al. \cite{tong2020digraph} leverage Perron--Frobenius theorem and the stationary distribution of strongly connected graphs to generalize the Laplacians of undirected graphs to directed graphs. They also eliminate the directed structure to guarantee that the normalized Laplacian is symmetric. Tong et al. \cite{tong2020digraph} further exploit the $k$th-order proximity between two nodes in a digraph by the inception network \cite{szegedy2016rethinking}, which not only allows the model to learn features of different sizes within one convolutional layer but also obtains larger receptive fields.
% However, transforming digraphs to undirected will not only mislead message passing scheme to aggregate the features with incorrect weights but also discard distinctive direction structure \cite{wang2020nodeaug}, such as irreversible time-series relationships. 
% For example, in a citation graph, a later published article can cite a previously published article, but not vice versa. This is a unique time-series relationship. If we convert it to an undirected graph, we lose this part of the information. Although we can represent the original directed graph in another form, such as a temporal graph learned by a combination of Recurrent Neural Networks (RNN) and GCN, we still want to mine more structural features from the original directed graph without adding additional components.
% Therefore, how to extend GCNs to the directed graphs needs to be explored.

There are several works that learn the specific directed structure.
In \cite{shi2019skeleton} and \cite{fu2020deep}, the incident matrix is used to combine the attributes of the incoming edges, outgoing edges and vertex itself to update the vertex and combine the attributes of the source nodes, target nodes and edge itself to update the edge.
Wang et al. \cite{wang2021multi} incorporate an attention mechanism to obtain the edge attention matrix or 1-hop attention matrix and further compute the attention scores of multi-hop neighbors via graph diffusion based on the powers of the 1-hop attention matrix and then define the graph attention diffusion based feature aggregation.
Zhang et al. \cite{zhang2021magnet} encode the undirected geometric structure in the magnitude of its entries and directional information in the phase of its entries based on a complex Hermitian matrix known as the magnetic Laplacian.
Thost and Chen \cite{thost2021directed} use attention mechanism to only aggregate directed predecessors information from the current layer and GRU to combine the neighbor and previous representation.
Beani et al. \cite{beani2021directional} exploit vector flows over graphs to develop a globally consistent directional and asymmetric aggregation function. They define a vector field in the graph by use of the Laplacian eigenvectors and develop a method of applying directional derivatives and smoothing by projecting node-specific messages into the field.

\subsection{Over-Smoothing}

From the perspective of information propagation, GNNs can be understood as an aggregation of information about neighboring nodes, and then after several GNNs messaging iterations, the representation of all nodes in the graph may become similar. This tendency is particularly common in basic GNN models and some models using self-loop update methods \cite{hamilton2020graph}. From the perspective of random wandering, when we use a $K$-layer GNN model, the effect of node $u$ on node $v$ is proportional to the probability of reaching node $v$ with $K$ random wandering steps starting from node $u$ \cite{xu2018representation}. However, when $K$ is large enough, the impact of each node approaches a smooth distribution of random wanderings on the graph, which means that local information is lost. The embedding of the nodes becomes over-smoothed, approaching an almost uniform distribution \cite{hoory2006expander}. The over-smoothing problem makes it impossible to construct deeper GNN models to exploit long-term dependencies in the graph, as these deep GNN models tend to generate only over-smoothed node embeddings. Some recent works aim to solve this problem. 
Xu et al. \cite{xu2018representation} perform concatenation, max-pooling or LSTM-attention operations on the output of each layer to adapt the receptive filed of each node instead of using only the output of the last layer. 
Klicpera, Bojchevski, and G{\"{u}}nnemann \cite{klicpera2019predict} use a variant of PageRank, personalized PageRank, considers the root node and preserve the information of the original features to some extent, whose update process is similar to the linear combination of the convolved features with the initial features. 
Pham et al. \cite{pham2017column} and Chen et al. \cite{chen2020simple} use an approach similar to the residual connectivity in convolutional neural networks (CNNs) to train deeper networks, i.e., the feature representation of a node is a linear interpolation between the previous representation of the node and the representation of neighboring nodes. 
Li et al. \cite{li2016gated} and Selsam et al. \cite{selsam2019learning} use the gated recurrent unit (GRU) and long short-term memory (LSTM) architecture to update the feature representation of the nodes. These strategies are very effective in facilitating deep GNN architectures (e.g., more than 10 layers) and preventing over-smoothing. 
Rong et al. \cite{rong2019dropedge} and Feng et al. \cite{NIPS2020_fb4c835f} use DropEdge and DropNode strategies and a framework for contrast learning, respectively, to address the over-smoothing problem.
However, these methods are proposed to address the over-smoothing problem for undirected graph, and there are few approaches for the directed graph.

\section{Preliminaries}
Let $\mathcal{G = (V, E)}$ be an undirected graph with $n$ nodes, where $\mathcal{V} = \{ 1, \dots, n \}$ is the set of $n$ nodes, $\mathcal{E \subseteq V \times V}$ is the set of edges connecting paired nodes in $\mathcal{V}$. The graph structure is represented by the $n \times n$ symmetric adjacency matrix $\mathbf{A} = \{ a_{ij} \}_{i,j=1}^n$, where $a_{ij} = 1$ if there exists an edge between nodes $i$ and $j$, and $a_{ij} = 0$ otherwise. The degree matrix $\mathbf{D} \in \mathbb{R}^{n \times n}$ is a diagonal matrix, where $\mathbf{D}_{ii} = \sum_i a_{ij} = \sum_j a_{ij}$. Let
$\mathbf{X} \in \mathbb{R}^{n \times d}$ denote the feature matrix of the $n$ nodes if graph $\mathcal{G}$ is associated with attributes. 
% \yincomment{with $d$ attributes?}
% \tancomment{each node is represented as a $d$-dimensional vector}
% \yincomment{Is the $d$-dimensional vector corresponding to $d$ attributes? Do attributes mean features?}
% \tancomment{Yes. For example, in citation network, nodes represent documents. Each document is encoded to a $d$-dimensional vector.}
If the node attributes are not given, we can encode nodes as a one-hot vector, that is, $\mathbf{X} = \mathbf{I}$, where $\mathbf{I} \in \mathbb{R}^{n \times n}$ is an identity matrix.
GNNs learn node or edge representation vectors by using the graph structure $\mathbf{A}$ and node feature matrix $\mathbf{X}$.
To facilitate subsequent discussions, we formalize an $L$-layers GNN under the architecture of MPNNs \cite{gilmer2017neural}, which computes the $l$-th layer representation vector $\mathbf{h}_v^l$ for node $v$ in a graph $\mathcal{G}$ as
\begin{equation*}
    \begin{aligned}
        \mathbf{m}_v^l &= \text{AGGREGATE}^l\left(\left\{ \mathbf{h}_u^{l-1} \ | \ u \in \mathcal{N}(v) \right\}\right)\\
        \mathbf{h}_v^l &= \text{COMBINE}^l\left(\mathbf{h}_v^{l-1}, \mathbf{m}_v^l\right), \ \ l = 1, \dots, L,
    \end{aligned}
\end{equation*}
where $\mathbf{h}_u^0 = \mathbf{X}_u$ is the input feature of node $u$, $\mathcal{N}(v)$ is a set of neighbors adjacent to $v$ in the graph. 
Under the framework of MPNNs, messages flow from $\mathcal{N}(v)$ to $v$ using an aggregation function (AGGREGATE), and then $\mathbf{h}_v^l$ of node $v$ is updated based on the corresponding neighbor messages $\mathbf{m}_v^l$ from the previous message aggregation step.

Many existing methods \cite{kipf2017semi, li2016gated, veli2018graph, xu2019powerful} are  specific examples %instantiation 
of the MPNNs framework and the differences lie in the choices of $\text{AGGREGATE}$ and $\text{COMBINE}$ functions. 
%The following are some well-known examples.

\subsection{GCN}

In the graph convolutional networks (GCN) \cite{kipf2017semi}, initially motivated by spectral graph convolutions \cite{bruna2014spectral, defferrard2016convolutional}, the $\text{AGGREGATE}$ and $\text{COMBINE}$ steps are integrated as follows,
\begin{equation*}
    \mathbf{h}_v^l = \text{ReLU}\left( \mathbf{W} \cdot \text{MEAN}\left\{ \mathbf{h}_u^{l-1}, \forall u \in \mathcal{N}(v) \cup \{ v \} \right\} \right),
\end{equation*}
where $\mathbf{W}$ is a linear parameter matrix and $\text{MEAN}$ represents an element-wise \textit{mean} pooling. 

\subsection{GraphSAGE} 
In the pooling variant of GraphSAGE \cite{hamilton2017inductive}, $\text{AGGREGATE}$ and $\text{COMBINE}$ are formulated as
\begin{equation*}
    \begin{aligned}
    \mathbf{m}_v^l &= \text{MAX}\left(\left\{ \text{ReLU}(\mathbf{W} \cdot \mathbf{h}_u^{l-1}), \forall u \in \mathcal{N}(v) \right\}\right) \\
    \mathbf{h}_v^l &= \mathbf{W} \cdot \left( \mathbf{h}_v^{l-1} \parallel \mathbf{m}_v^l \right),
    \end{aligned}
\end{equation*}
where $\text{MAX}$ represents an element-wise \textit{max} pooling and $\parallel$ denotes the concatenation operation.
% The COMBINE step could be a concatenation followed by a linear mapping $\mathbf{W} \cdot (h_v^{l-1} \parallel m_v^l)$ as in GraphSAGE.

For node-level tasks, the node representation $\mathbf{h}_v^L$ of the final layer $L$ is used for prediction. For graph-level tasks, the $\text{READOUT}$ function computes a feature vector for the whole graph using some readout function according to
\begin{equation*}
    \mathbf{h}_{\mathcal{G}} = \text{READOUT}\left( \left\{ h_v^L, v \in \mathcal{V} \right\} \right),
\end{equation*}
where
$\text{READOUT}$ can be a simple permutation invariant function such as summation or a more sophisticated graph-level pooling function \cite{ying2018hierarchical, zhang2018end}.

\section{Method}

\begin{figure}[t]
    \centering
    \includegraphics[width=1.0\columnwidth]{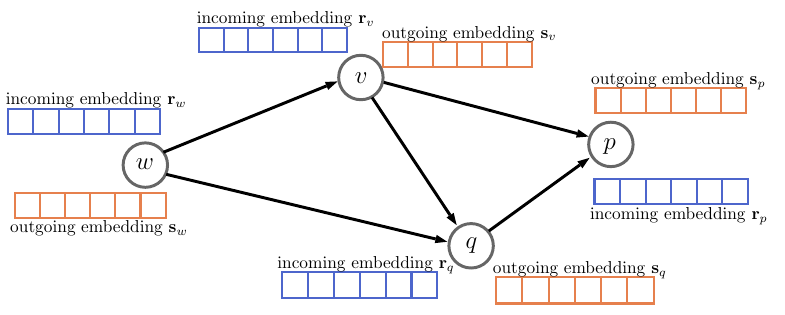}
    \caption{Illustration of the outgoing embedding and the incoming embedding over a four-node toy graph. For example, node $v$ has both the outgoing embedding $\mathbf{s}_v$ and incoming embedding $\mathbf{r}_v$, which describe the sending and receiving pattern respectively. Node $w$ has two outgoing edges but no incoming edge, hence its incoming embedding $\mathbf{r}_w=\mathbf{0}$. Likewise, the outgoing embedding $\mathbf{s}_p=\mathbf{0}$ because node $p$ only contains incoming edges. The occurrence of a directional edge can be interpreted by the two types of embedding. For instance, the likelihood of edge $w\to v$ depends on $\mathbf{s}_w$ and $\mathbf{r}_v$. 
    }
    \label{fig:inoutEmbedding}
\end{figure}

Traditionally, in undirected graphs, relationships are symmetric, and thus a node is represented by a single vector embedding $\mathbf{h}$. Directed graphs, in contrast, have asymmetric node relationships. In particular, some nodes can have edges of both in-taking (receiving information) and  out-flowing (sending information) functions, which would make GNNs learn the graph representation more difficult than an undirected setting. To capture the asymmetries, we propose to represent node $v$ in a directed graph with an outgoing (sending information) embedding vector $\mathbf{s}_v$ and an incoming (receiving information) embedding vector $\mathbf{r}_v$, as shown in Figure \ref{fig:inoutEmbedding}. Based on the two different roles of each node, we develop an asymmetric GNN, which contains two steps with the first one to aggregate/update its incoming features and the second one to aggregate/update its outgoing features.

Formally, given a directed graph $\mathcal{G} = (\mathcal{V}, \mathcal{E})$ with $n$ nodes, we have an asymmetric adjacency matrix $\mathbf{A} = \{ a_{ij} \}_{i,j=1}^n$, where $a_{ij} = 1$ if there exists a directed edge pointing from node $i$ to node $j$, and $a_{ij} = 0$ otherwise. Specifically, each node can be split into an out-node and an in-node, and the rows of $\mathbf{A}$ index the out-nodes and the columns of $\mathbf{A}$ index the in-nodes. Thus, the directed graph has two diagonal degree matrices: an in-degree matrix $\mathbf{P} \in \mathbb{R}^{n \times n}$ with $\mathbf{P}_{ii} = \sum_i^n a_{ji}$ and an out-degree matrix $\mathbf{O} \in \mathbb{R}^{n \times n}$ with $\mathbf{O}_{ii} = \sum_j^n a_{ij}$. Note that the in-degree and out degree of each node are equal in undirected graph, because of the symmetrical $\mathbf{A}$. The nodes are characterized by the feature matrix $\mathbf{X} \in \mathbb{R}^{n \times d}$ which is the same as undirected graphs. 

The main idea of asymmetric graph neural network (AGNN) is to process nodes according to the different behaviors in terms of sending and receiving patterns so as to learn receiving and sending feature maps, respectively.

\subsection{Asymmetric Message Passing}
We develop a multi-layer AGNN with the following rules,
\begin{equation}
    \begin{aligned}
        \mathbf{m}_v^l &= \text{AGGREGATE}_1^l\left(\left\{ \mathbf{r}_u^{l-1} \ | \ u \in \mathcal{N}_{in}(v) \right\}\right)\\
        \mathbf{a}_v^l &= \text{AGGREGATE}_2^l\left(\left\{ \mathbf{s}_w^{l-1} \ | \ w \in \mathcal{N}_{out}(v) \right\}\right)\\
        \mathbf{r}_v^l &= \text{COMBINE}_1^l\left(\mathbf{r}_v^{l-1}, \mathbf{m}_v^l\right)\\
        \mathbf{s}_v^l &= \text{COMBINE}_2^l\left(\mathbf{s}_v^{l-1}, \mathbf{a}_v^l\right),
    \end{aligned}
\label{eq:1}
\end{equation}
where $\mathbf{r}_v^0 = \mathbf{s}_v^0 = \mathbf{X}_v$ is the input feature of $v$, $\mathcal{N}_{in}(v)$ and $\mathcal{N}_{out}(v)$ denote the sets of direct predecessors and successors of $v$ respectively. More specifically, for $\forall u \in \mathcal{V}$, if there exists an edge $u \to v$, then $u \in \mathcal{N}_{in}(v)$; likewise, for $\forall w \in \mathcal{V}$, if there exists an edge $v \to w$, then $w \in \mathcal{N}_{out}(v)$. $l = 1, 2, \dots, L$ denotes the layer of convolution. Finally, we can model the different roles of receiving and sending information, and obtain two embeddings of one node (i.e., the receiving messages flow from $\mathcal{N}_{in}(v)$ to the incoming embedding $\mathbf{r}_v$ and the sending messages flow from $\mathcal{N}_{out}(v)$ to the outgoing embedding $\mathbf{s}_v$ for node $v$, respectively). 

For node-level tasks, the two node representations $\mathbf{r}_v^L$ and $\mathbf{s}_v^L$ of the final layer are merged and used for prediction,
\begin{equation}
    \mathbf{z}_{v} = \text{COMBINE}_3\left( \mathbf{r}_v^L, \mathbf{s}_v^L \right),
    \label{eq:2}
\end{equation}
where $L$ is the number of layers.
For graph-level tasks, the readout phase computes a feature vector for the whole graph according to
\begin{equation*}
    \mathbf{z}_{\mathcal{G}} = \text{READOUT}\left( \{ \mathbf{z}_v, v \in \mathcal{V} \} \right).
\end{equation*}

By defining the specific $\text{AGGREGATE}$ and $\text{COMBINE}$ functions, we would obtain a corresponding directed GNN method.
For instance, consider the element-wise \textit{mean} pooling in the framework of the asymmetric message passing, it is a directed version of GCN,
\begin{equation}
    \begin{aligned}
            \mathbf{S}^l &= \sigma \left( \tilde{\mathbf{A}} \mathbf{S}^{l-1} \mathbf{W}_1^l \right)\\
            \mathbf{R}^l &= \sigma \left( \hat{\mathbf{A}} \mathbf{R}^{l-1} \mathbf{W}_2^l \right),
    \end{aligned}
\label{eq:3}
\end{equation}
where $\mathbf{S}^0 = \mathbf{R}^0 = \mathbf{X} \in \mathbf{R}^{n \times d}$, $\sigma$ is an activation function (e.g., $\text{ReLU}$), $\tilde{\mathbf{A}} = \mathbf{O}^{-1/2} (\mathbf{A} + \mathbf{I}) \mathbf{P}^{-1/2} \in \mathbb{R}^{n \times n}$ is the normalized adjacency matrix of the directed graph $\mathcal{G}$ with added self-connections, $\hat{\mathbf{A}} = \mathbf{P}^{-1/2} (\mathbf{A}^T + \mathbf{I}) \mathbf{O}^{-1/2} \in \mathbb{R}^{n \times n}$ and $\mathbf{W}_1^{l}, \mathbf{W}_2^{l} \in \mathbb{R}^{d_{l-1} \times d_l}$ are trainable parameters. 
$\mathbf{S}^l$ and $\mathbf{R}^l$ are the sending and reveiving embedding of the $l$-th layer in matrix form.
Through Eq. (\ref{eq:3}), we can aggregate (i.e., \textit{mean}) the predecessor (or successor) and root nodes to matrix $\mathbf{R}^l$ (or $\mathbf{S}^l$). We let $\mathbf{S}:=\mathbf{S}^L$ and $\mathbf{R}:=\mathbf{R}^L$ for notation simplicity, where $\mathbf{S}^L$ and $\mathbf{R}^L$ are the outgoing and incoming embeddings of the $L$-th layer.

For node classification, we can coalesce the incoming and outgoing embedding vectors, followed by a softmax function to map every node to a probability distribution,
\begin{equation*}
    \mathbf{Z} = \text{Softmax}\left( \mathbf{R} + \mathbf{S} \right).
\end{equation*}

For graph classification, we can concatenate the two embeddings, and a \textit{sum} pooling is followed to obtain the graph-level representation,
\begin{equation*}
    \begin{aligned}
        \mathbf{Z} &= \text{FC} \left(\mathbf{R} \parallel \mathbf{S} \right) \in \mathbb{R}^{n \times d}, \\
        \mathbf{Z}_{\mathcal{G}} &= \text{SUM}\left( \mathbf{Z} \right) \in \mathbb{R}^{1 \times d},
    \end{aligned}
\end{equation*}
where $\text{FC}$ denotes the fully-connected layer and $\parallel$ denotes the concatenate operation. 

\subsection{Loss Function}

\subsubsection{Classification Loss}

For semi-supervised multi-class classification, we evaluate the cross-entropy error over all labeled examples,
\begin{equation*}
    \mathcal{L}_{\text{error}} = - \sum_{l \in \mathcal{Y}_L} \sum_{c=1}^C \mathbf{Y}_{lc} \ln \mathbf{Z}_{lc},
\end{equation*}
where $\mathcal{Y}_L$ is the set of node (graph) indices that have labels, $\mathbf{Y}_{lc}$ is the true label, $\mathbf{Z}_{lc}$ is the predicted label and $C$ is the number of classes.

\subsubsection{Regularization Formulation.} \label{sec:4.2.2}
With the proposed asymmetric node embeddings for directed graphs, the likelihood of a directional edge can be calculated based on the outgoing embedding of its source node and the incoming embedding of its target node. As shown by the example in Figure \ref{fig:inoutEmbedding}, the occurrence of the edge $v \to q$ can be interpreted together by the vectors $\mathbf{s}_v$ and $\mathbf{r}_q$. 
The likelihood of graph $\mathcal{G}$ can be jointly calculated based on the likelihood of edges. A regularization formulation equips the log-likelihood of $\mathcal{G}$ to avoid over-smoothing for the proposed model. As the number of layers increases, the embeddings of nodes may be similar due to lager receptive fields through message passing, while the regularization formulation would maintain the virginity that nodes are similar to their direct neighborhoods. In other words, nodes connected by the same edge preserve similar features.

In particular, considering the outgoing embedding set $\{ \mathbf{s}_i \}_{i=1}^n$ and the incoming embedding set $\{ \mathbf{r}_j \}_{j=1}^n$ for all the nodes, we formulate the likelihood of an edge $i \to j$ with the inner product between $\mathbf{s}_i$ and $\mathbf{r}_j$ followed by a sigmoid function,
% In the dimension reduction space, the structure and inherent properties of the directed graph can be preserved, so that the embedding vectors of similar nodes are close as well. Consequently, the community structures of the  sending and receiving information can be modeled through some distance measures among $\{ \mathbf{s}_i \}_{i=1}^n$ and $\{ \mathbf{r}_j \}_{j=1}^n$. We consider the inner product (cosine-similarity) of the two embeddings followed by a sigmoid function to measure the similarity of two nodes,
\begin{equation*}
    p_{ij} = \text{sigmoid}\left(\mathbf{s}_i^T \mathbf{r}_j \right) = \frac{1}{1 + \exp(-\mathbf{s}_i^T \mathbf{r}_j)},
\end{equation*}
where $p_{ij} = P(a_{ij} = 1)$, and $\mathbf{s}_i$ is the outgoing embedding of node $i$ and $\mathbf{r}_j$ is the incoming embedding of node $j$ obtained from Eq. (\ref{eq:1}).

Consequently, the likelihood function of $\mathcal{G}$ is
\begin{equation*}
    P(\mathcal{G}) = \prod_{i,j=1}^n p_{ij}^{a_{ij}} (1-p_{ij})^{1-a_{ij}} = \prod_{i,j=1}^n \frac{\exp (\mathbf{s}_i^T \mathbf{r}_j a_{ij})}{1+\exp(\mathbf{s}_i^T \mathbf{r}_j)}.
\end{equation*}
This leads to the negative log-likelihood function (regularization),
\begin{equation*}
    \begin{aligned}            \mathcal{L}_{\text{reg}}(\mathbf{s}, \mathbf{r}) 
            &= - \frac{1}{n^2} \sum_{i,j=1}^n \Big(\mathbf{s}_i^T \mathbf{r}_j a_{ij} - \log\big(1 + \exp(\mathbf{s}_i^T \mathbf{r}_j) \big) \Big)\\
            &= - \frac{1}{n^2} \mathbf{1}^T \Big(\mathbf{S}^T \mathbf{R} \odot \mathbf{A} - \log\big(1 + \exp(\mathbf{S}^T \mathbf{R}) \big) \Big) \mathbf{1},
    \end{aligned}
\end{equation*}
where $\mathbf{1}$ denotes the vector with all entries being $1$, $\mathbf{S}$ and $\mathbf{R}$ are given in Eq. (3) and $\odot$ denotes the Hadamard product.
%\yincomment{Here, $\mathbf{S}$ $\mathbf{R}$ different from those in Eq. (2)? If so, we may say they are the matrix forms.}
%\liucomment{$\mathbf{S}$ here should be $\mathbf{S}^L$, we let $S:=\mathbf{S}^L$,$R:=\mathbf{R}^L$ for simple formulation. In Eq 3 and 4, the subscript $L$ have been removed to unified the formulation.}
Finally, we obtain the total loss function,
\begin{equation}
    \mathcal{L} = \mathcal{L}_{\text{error}} + \lambda \mathcal{L}_{\text{reg}},
\label{eq:4}
\end{equation}
where $\lambda$ is a hyper-parameter as the regularization coefficient, serving as a trade-off between the classification error and regularization.

Furthermore, the proposed method is very flexible and can be extend to undirected graph by setting $\mathbf{r}_v = \mathbf{s}_v$ for each node $v$. 

\section{Experiment}

We evaluate the effectiveness of our model on both node-level and graph-level tasks. Node-level experiments are performed on citation and co-purchase datasets, graph-level on a neural architecture dataset. A common task of GNNs is semi-supervised node classification, which can be used to verify the learning capability of our model. Another task is graph regression, which can be used to verify that our model is not node-level-specific and can be easily adapted to graph-level task. Compared with the canonical experiments for undirected graphs, the main difference lies in that the task under our setup is applied to the directed graphs and the given adjacency matrix $\mathbf{A}$ is asymmetric. The Supplementary Material reports details on the experiments and reproducibility.

\subsection{Experimental Settings}
\subsubsection{Datasets.} For semi-supervised node classification task, we apply the AGNN on four directed graph datasets with detailed information given in Table \ref{dataset_detail}. The four real datasets include two citation datasets: Cora-ML \cite{bojchevski2018coraml} and Citeseer \cite{sen2008collective}, and two Amazon co-purchase datasets \cite{shchur2018amazon}: AM-Computer and AM-Photo. In our experiments, we randomly split the datasets and perform multiple experiments for obtaining stable and reliable results. The mean and standard deviation of classification accuracy are calculated by running each experiment $20$ times with random weight initialization. For training/validation/testing split, we follow the rules in DiGCN \cite{tong2020digraph} to choose 20 labels per class for the training set, 500 labels for the validation set and the rest for the testing set. For graph regression task, we apply our model on NA dataset \cite{zhang2019na}. The NA dataset contains 19020 neural architectures from the ENAS software. The task is to evaluate each neural architecture’s weight-sharing accuracy on CIFAR-10 \footnote{An image classification dataset. \url{http://www.cs.toronto.edu/~kriz/cifar.html}}. Since it is a regression task, we use the RMSE, MAE and MAPE metrics. Followed by \cite{thost2021directed}, we randomly split the dataset into 90\% training and 10\% held-out test sets. 

\subsubsection{Baselines.} We compare our model with seven state-of-the-art models that can be divided into three main categories: (i) spectral-based GNNs including GCN \cite{kipf2017semi}, SGC \cite{wu2019simplifying}, APPNP \cite{klicpera2019predict}; (ii) spatial-based GNNs including GraphSage \cite{hamilton2017inductive} and GAT \cite{veli2018graph}; and (iii) directed GNNs including DGCN \cite{tong2020directed} and DiGCN \cite{tong2020digraph}.

\subsubsection{Implementation Details.} For citation datasets, Cora-ML and Citeseer, we train our model for a maximum of 1000 epochs with early stopping if the validation accuracy does not increase for 200 consecutive epochs and the hidden layer dimension is 64. For Amazon co-purchase datasets, AM-Computer and AM-Photo, we train our model for a maximum of 500 epochs without early stopping and the hidden layer dimension is 128. In the four node-level datasets, the learning rate is 0.01, the weight decay is 0.0005, and the dropout rate is 0.5. We calculate the mean test accuracy with standard deviation in percent (\%) averaged over 20 random dataset splits with random weight initialization. For NA dataset, we train our model for a maximum of 100 epochs and the hidden layer dimension is 64. The learning rate is 0.001, the batch size is 32 and the dropout rate is 0.5. 
% More details about the implementation can be found in the supplementary materials. 

\begin{table}[t]
\centering
\resizebox{1.\columnwidth}{!}{
\begin{tabular}{lccccc}
\toprule
Datasets    & Nodes & Edges  & Classes & Features & Label rate    \\ \midrule
% Cora        & 19793 & 65311  & 70      & 8710     & 7.07\%        \\
Cora-ML                         & 2995  & 8416   & 7       & 2879     & 4.67\%        \\
Citeseer                        & 3312  & 4715   & 6       & 3703     & 3.62\%        \\
Am-Computer                     & 13752 & 287209 & 10      & 767      & 1.45\%        \\
Am-Photo                        & 7650  & 143663 & 8       & 745      & 2.10\%        \\ \bottomrule
\end{tabular}
}
\caption{Detailed information on the four real datasets. We choose 20 labels per class for the training set following DiGCN \cite{tong2020digraph}.}
\label{dataset_detail}
\end{table}

\subsection{Experimental Results}

% \begin{figure}[t]
%     \centering
%     \includegraphics[width=1.0\columnwidth]{Figure/acc_layer.pdf}
%     \caption{Influence of the model depth (number of layers) on classification performance. Markers denote mean classification accuracy (training vs. testing) for 20 random dataset splits. Shaded areas denote the standard error. We show results for our AGNN model with regularization as well as those without regularization.}
%     \label{fig:acc_layer}
% \end{figure}

\begin{table*}[t]
\centering
\begin{tabular}{lcccc}
\toprule
Models                                           & Cora-ML                        & Citeseer                        & Am-Computer                      & Am-Photo                  \\ \midrule
GCN                                              & 53.11 ± 0.8                    & 54.36 ± 0.5                     & 60.50 ± 1.6                      & 53.20 ± 0.4               \\ 
SGC                                              & 51.14 ± 0.6                    & 44.07 ± 3.5                     & 76.17 ± 0.1                      & 71.25 ± 1.3               \\
APPNP                                            & 70.07 ± 1.1                    & 65.39 ± 0.9                     & 63.16 ± 1.4                      & 79.37 ± 0.9               \\
GraphSage                                        & 72.06 ± 0.9                    & 63.19 ± 0.7                     & 79.29 ± 1.3                      & 87.57 ± 0.9               \\
GAT                                              & 71.91 ± 0.9                    & 63.03 ± 0.6                     & 79.45 ± 1.5                      & 89.10 ± 0.7               \\
DGCN                                             & 75.02 ± 0.5                    & 66.00 ± 0.4                     & /                                & 83.66 ± 0.8               \\
DiGCN                                            & 80.28 ± 0.5                    & 66.11 ± 0.7                     & \textbf{85.94 ± 0.5}             & 90.02 ± 0.5               \\ \midrule
%Ours                                                                    \\
AGNN w/o reg                                     & \it{80.52 ± 0.3}               & \it{69.07 ± 0.3}                & 84.27 ± 0.7                      & \it{90.67 ± 0.3 }         \\
AGNN w reg                                       & \textbf{80.76 ± 0.2}           & \textbf{69.21 ± 0.3}            & \it{84.34 ± 0.7}                 & \textbf{90.70 ± 0.3}      \\ \bottomrule
\end{tabular}
\caption{Overall accuracy comparison on node classification between our AGNN without regularization and AGNN with regularization and seven existing methods. The best results are highlighted in \textbf{boldface} and the second in \it{Italian} font.}
\label{table:overall_1}
\end{table*}

\subsubsection{Comparisons with the state-of-the-art methods on node classification.} 
It can be seen from Table \ref{table:overall_1} that our method achieves the state-of-the-art classification accuracy on citation datasets and comparable results on Amazon co-purchase datasets. 
The spectral-based models, including GCN and SGC, perform poorly on directed graph datasets in contrast to their better performances in undirected graphs. The possible explanation is that these models are limited to aggregate features from the direct successors using asymmetric adjacency matrices so that they can only capture the sending information but not the receiving information. 
APPNP is an exception because it allows features to be propagated randomly with a certain teleport probability, which breaks the path restriction and achieves good performance in directed graphs.
Both GraphSage and GAT are spatial-based methods, which yield better performances than APPNP, demonstrating that these methods are more suitable for directed graphs. GraphSage generates embeddings by sampling and aggregating features from nodes' local neighborhoods, which can capture more information than direct successors. GAT specifies different weights to different neighborhoods, which reflects differences in the edges. DGCN performs well on three datasets, while it leads to out of memory on AM-Computer because the method uses both the first-  and second-order proximity matrices to obtain structural features in directed graphs. DiGCN performs better than our model only on the Amazon-Computer dataset, may be due to the use of inception method. However, as shown by the results on the other three datasets, our model performs better than all the existing methods. 
%which verifies our idea. 
The incoming embedding and outgoing embedding can effectively accommodate the asymmetric structure of directed graphs. Moreover, the accuracy of our model improves significantly when regularization is incorporated, which is consistent with the analysis in Section 4.2.2.

\subsubsection{Comparisons with the state-of-the-art methods on graph-level task.} Table \ref{table:overall_2} shows that AGNN outperforms the six baselines on graph regression task. While GCN and GraphSage achieve work well, AGNN proves the superiority of capturing directed structure. Note that we use same readout function (Sum operation) in these experiments.

\begin{table}[t]
\centering
\begin{tabular}{lccc}
\toprule
Model          & RMSE              & MAE              & MAPE   \\ \midrule
GCN            & \it{0.0051}       & \it{0.0040}      & \it{0.0055} \\
SGC            & 0.0100            & 0.0081           & 0.0110 \\
APPNP          & 0.0060            & 0.0046           & 0.0063 \\
GraphSage      & 0.0053            & 0.0041           & 0.0056 \\
GAT            & 0.0055            & 0.0042           & 0.0057 \\
DiGCN          & 0.0299            & 0.0293           & 0.0397 \\ \midrule
AGNN           & \textbf{0.0050}   & \textbf{0.0039}  & \textbf{0.0053} \\ \bottomrule
\end{tabular}
\caption{Accuracy comparison on graph regression task between our AGNN and six existing methods. The best results are highlighted in \textbf{boldface} and the second in \it{Italian} font.}
\label{table:overall_2}
\end{table}

\subsection{Ablation Study}

\subsubsection{Accuracy with different regularization coefficients.} 
\begin{figure}[t]
    \centering
    \includegraphics[width=0.95\columnwidth]{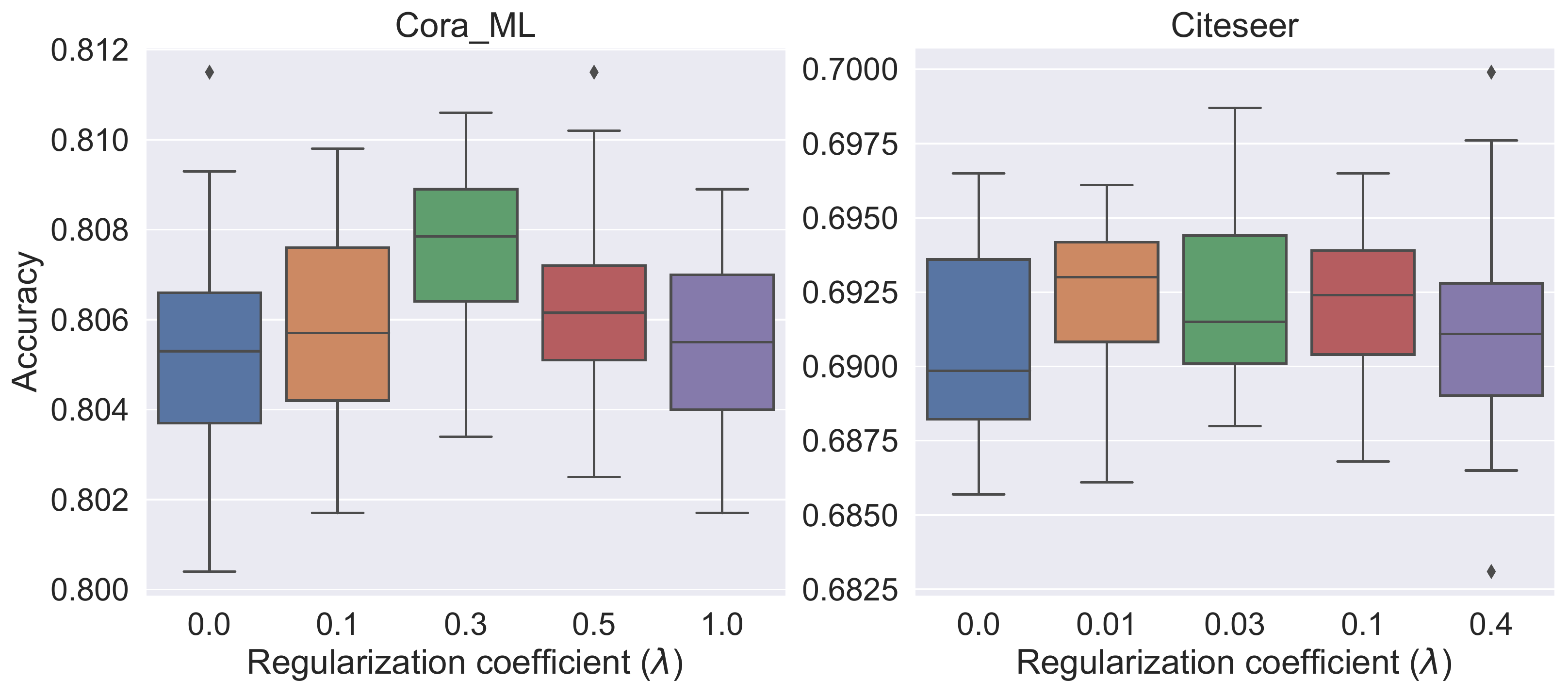}
    \caption{Influence of the regularization  coefficient $\lambda$ on classification performance.}
    \label{fig:coef}
\end{figure}

We validate our model with different values of the regularization coefficient $\lambda$ in Eq. (\ref{eq:4}). As shown in Figure \ref{fig:coef},
the models with regularization perform better than without ones in the Cora-ML and Citeseer datasets, which proves the superiority of the regularization term. The value of the regularization coefficient should be adjusted for different datasets, because different graphs exhibit different neighborhood structures \cite{klicpera2019predict}. We choose $\lambda$ according to the mean test accuracy of 20 times experiments. The detailed descriptions are summarized in the Supplementary Material.

\subsubsection{Accuracy with different fusion functions.} 
\begin{figure}[t]
    \centering
    \includegraphics[width=0.95\columnwidth]{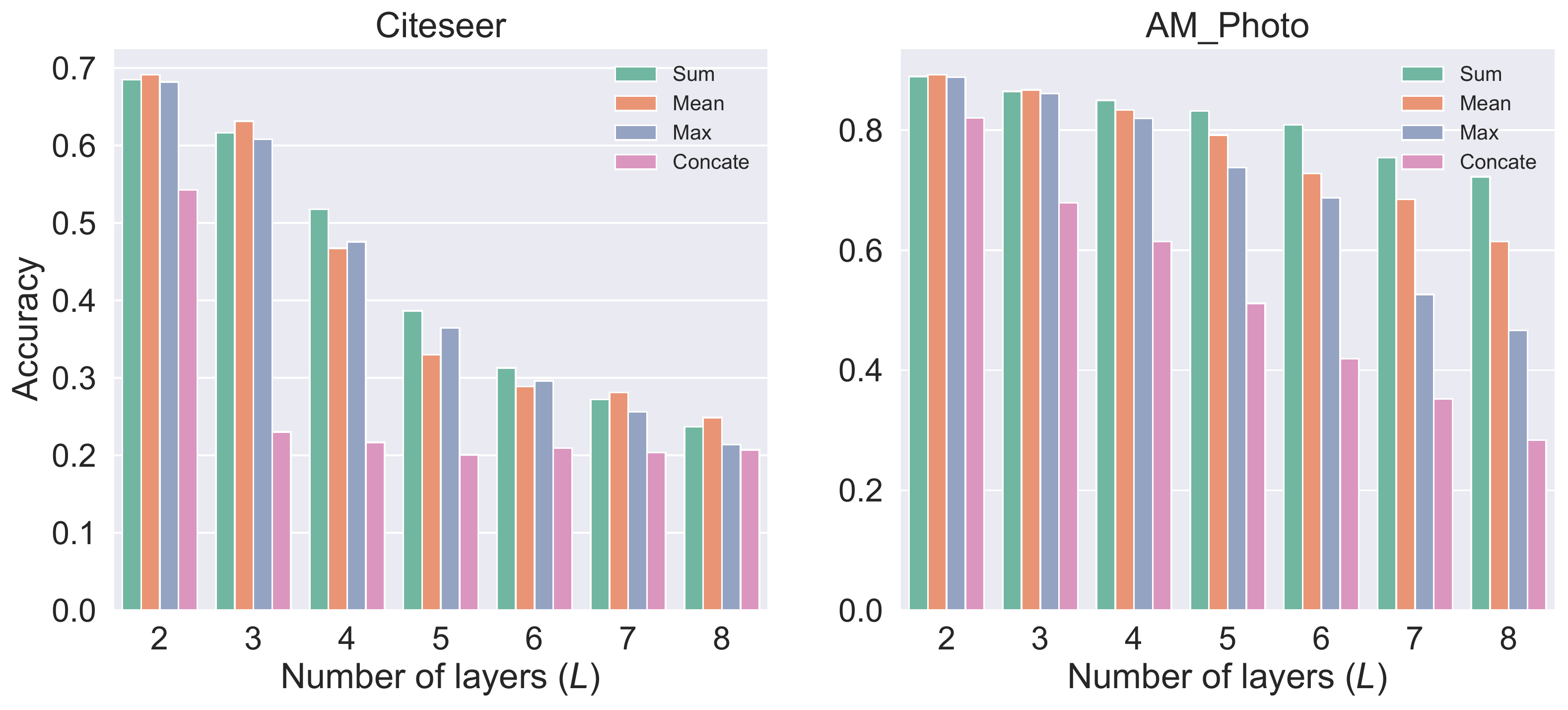}
    \caption{Influence of fusion functions on classification performance.}
    \label{fig:fusion}
\end{figure}

Figure \ref{fig:fusion} illustrates the results with different fusion operations (i.e. the $\text{COMBINE}_3$ function mentioned in Eq. (\ref{eq:2})), including Sum, Max, Mean, and Concatenate operations.
Sum and Mean achieve better results in proposed model because they require fewer parameters which can prevent the model from over-fitting and thus capture more information \cite{xu2019powerful}.

\section{Conclusion}

Most of the existing GNNs focus on undirected graphs, while the directed graphs are commonly encountered across different domains. The directional edge and asymmetric relationship are the most challenging aspects that make the directed GNNs difficult to train and implement in practice. We propose a simple yet effective method to model the asymmetric property of the directed graph within the GNN framework. The idea is to let each node play two asymmetric roles simultaneously: one is to act as an outgoing node, another is to act as an incoming node. The outgoing and incoming roles are quantified by two different embedding vectors. The outgoing embedding aims to capture the sending features of a node, and the incoming embedding aims to model the receiving features. With the two embeddings of each node, any directional edge in the graph can be interpreted by the outgoing embedding of its source node and the incoming embedding of its target node. Such one node but two roles setting can be easily applied to the current GNNs, by defining two aggregating/updating steps: one for aggregating/updating the receiving features of nodes, another for aggregating/updating the sending features of nodes. Experimental results on multiple real-world datasets illustrate that the proposed asymmetric GNN yields good performance in directed graph tasks.

%%%%%%%%% REFERENCES
{\small

\bibliographystyle{ieee_fullname}
}

\end{document}